\documentclass[lettersize,journal]{IEEEtran}
\usepackage{amsmath,amsfonts}
\usepackage{algorithmic}
\usepackage{algorithm}
\usepackage{array}
\usepackage[caption=false,font=normalsize,labelfont=sf,textfont=sf]{subfig}
\usepackage{textcomp}
\usepackage{stfloats}
\usepackage{url}
\usepackage{verbatim}
\usepackage{graphicx}
\usepackage{cite}
\usepackage{multirow}
\usepackage{booktabs}
\usepackage{color}
\begin{document}

\title{3D Weakly Supervised Semantic Segmentation via Class-Aware and Geometry-Guided Pseudo-Label Refinement}

\author{Xiaoxu~Xu,
        Xuexun Liu,
        Jinlong Li,
	Yitian~Yuan,
	Qiudan~Zhang,~\IEEEmembership{Member,~IEEE,}
	Lin~Ma,~\IEEEmembership{Senior Member,~IEEE,}
        Nicu Sebe,~\IEEEmembership{Senior Member,~IEEE,}
   	Xu~Wang,~\IEEEmembership{Member,~IEEE}
        % <-this % stops a space
\thanks{This work was supported in part by the National Natural Science Foundation of China under Grants 62371310, in part by the Guangdong Basic and Applied Basic Research Foundation under Grant 2023A1515011236, in part by the Stable Support Project of Shenzhen under Grant 20231122122722001. \textit{(Corresponding author: Qiudan Zhang.)}

Xiaoxu Xu is with the College of Computer Science, Beihang University, Beijing, 100191, China. E-mail: xuxiaoxu68@163.com.

Xuexun Liu, Qiudan Zhang and Xu Wang are with the College of Computer Science and Software Engineering, Shenzhen University, Shenzhen, 518060, China. Email: (2019043026@email.szu.edu.cn, qiudanzhang@szu.edu.cn and wangxu@szu.edu.cn and ).

Jinlong Li and Nicu Sebe are with the Department of Information Engineering and Computer Science, University of Trento, Trento, 38100, Italy. E-mail:  jinlong.li@unitn.it and  niculae.sebe@unitn.it.

Yitian Yuan and Lin Ma are with Meituan, Beijing, China. E-mail: yuanyitian@foxmail.com and forest.linma@gmail.com.
}}

% The paper headers
\markboth{Submitted to IEEE Transactions on Image Processing, July 2025}%
{Shell \MakeLowercase{\textit{et al.}}: A Sample Article Using IEEEtran.cls for IEEE Journals}

\maketitle

\begin{abstract}
3D weakly supervised semantic segmentation (3D WSSS) aims to achieve semantic segmentation by leveraging sparse or low-cost annotated data, significantly reducing reliance on dense point-wise annotations. Previous works mainly employ class activation maps or pre-trained vision-language models to address this challenge. However, the low quality of pseudo-labels and the insufficient exploitation of 3D geometric priors jointly create significant technical bottlenecks in developing high-performance 3D WSSS models. In this paper, we propose a simple yet effective 3D weakly supervised semantic segmentation method that integrates 3D geometric priors into a class-aware guidance mechanism to generate high-fidelity pseudo labels. Concretely, our designed methodology first employs Class-Aware Label Refinement module to generate more balanced and accurate pseudo labels for semantic categrories. This initial refinement stage focuses on enhancing label quality through category-specific optimization. Subsequently, the Geometry-Aware Label Refinement component is developed, which strategically integrates implicit 3D geometric constraints to effectively filter out low-confidence pseudo labels that fail to comply with geometric plausibility. Moreover, to address the challenge of extensive unlabeled regions, we propose a Label Update strategy that integrates Self-Training to propagate labels into these areas. This iterative process continuously enhances pseudo-label quality while expanding label coverage, ultimately fostering the development of high-performance 3D WSSS models. Comprehensive experimental validation reveals that our proposed methodology achieves state-of-the-art performance on both ScanNet and S3DIS benchmarks while demonstrating remarkable generalization capability in unsupervised settings, maintaining competitive accuracy through its robust design.
  
\end{abstract}


\section{Introduction}
\label{sec:intro}
\begin{figure}[t]
  \centering
   \includegraphics[width=1\linewidth]{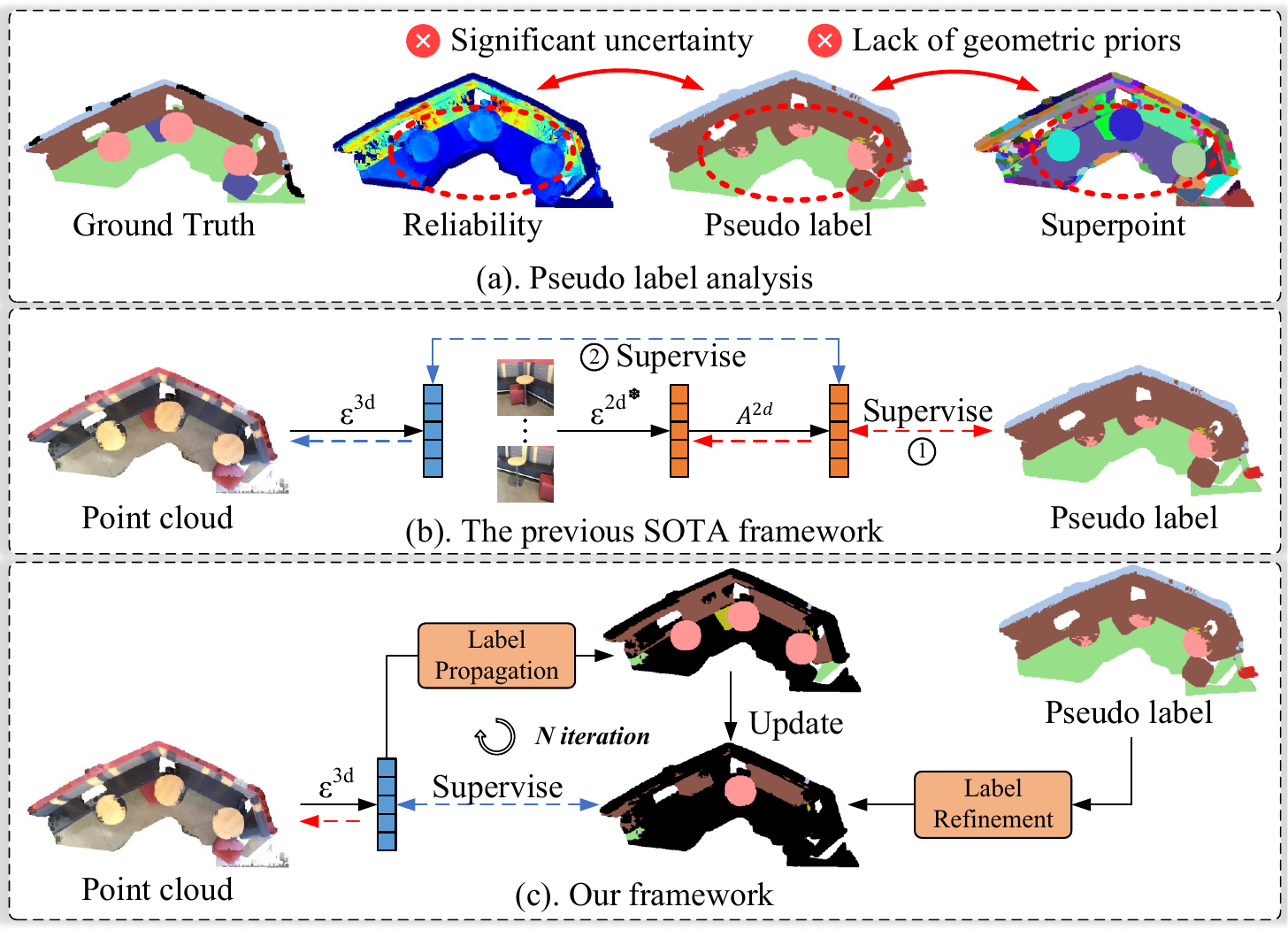}
   \caption{\small Comparison between previous 3D WSSS methods and our proposed approach. (a) Pseudo label analysis; (b) The typical pipeline of previous SOTA methods; (c) The workflow of our proposed methodology.}
   \label{fig1}
\end{figure}

\IEEEPARstart{P}{oint} cloud semantic segmentation~\cite{qi2017pointnet,yan20222dpass,qian2022pointnext} serves as a pivotal technique for jointly extracting geometric and semantic information from 3D scene data, attracting considerable attention in recent years. While fully supervised approaches have achieved remarkable performance, their reliance on labor-intensive point-level annotations remains a critical limitation. To alleviate this annotation burden, weakly supervised learning has emerged as a cost-effective alternative, utilizing less detailed supervision signals such as subcloud-level~\cite{wei2020multi} or scene-level labels~\cite{ren20213d,yang2022mil,kweon2022joint}. Among these, scene-level annotations offer particular advantages by providing holistic supervision over entire 3D scenes rather than dense per-point labels. Previous methods~\cite{choy20194d,yang20232d,ren20213d} for 3D weakly supervised semantic segmentation frequently adopt class activation maps~\cite{zhou2016learning} as a foundational technique when operating under scene-level supervision. More recently, vision-language models (VLMs)~\cite{ghiasi2022scaling} have been integrated into this domain, bridging 2D image understanding with 3D textual semantics. For instance, as illustrated in Fig.~\ref{fig1}(b), most of the existing methods typically follow a two-stage framework: (1) leveraging pretrained VLM~\cite{ghiasi2022scaling} to generate pseudo labels for refining 2D feature embeddings as indicated by the red dashed line, which are subsequently projected into 3D space; and (2) training a 3D network to exploit these refined embeddings for learning spatially aware representations, as indicated by the blue dashed line. 
% This hybrid paradigm combines the semantic richness of language models with the geometric fidelity of 3D point clouds, advancing the field toward more annotation-efficient solutions.

Although leveraging VLMs in 3D WSSS models demonstrates promising potential, several key challenges persist. As illustrated in Fig.~\ref{fig1}(a), we observe that most pseudo labels contain low probabilities and lack 3D geometric priors, those with correct predictions blur category boundaries, ultimately resulting in model performance degradation. Additionally, previous methods overly depend on pretrained VLMs, overlooking inherent 3D geometric priors. However, directly integrating explicit 3D geometric features, such as point normals, into networks is still challenging. Meanwhile, we find that unsupervised superpoint segmentation has shown potential in exploiting 3D geometric structures for benefiting segmentation~\cite{sun2023superpoint,landrieu2018large}. This motivates us to investigate whether superpoint-based representations, which implicitly encode 3D geometric priors, can improve segmentation performance while addressing the limitations of purely 2D VLM-driven approaches.

In this paper, we propose a simple yet effective weakly supervised approach to 3D semantic segmentation that operates through a two-stage paradigm as illustrated in Fig.~\ref{fig1}(c), to significantly improve the quality of pseudo labels. Specifically, in the first stage, we design a Pseudo Label Generation and Refinement procedure that produces high-quality point-level pseudo labels under 3D WSSS with only scene-level supervision. This procedure employs two key modules: Class-Aware Label Refinement (CALR) and Geometry-Aware Label Refinement (GALR). The CALR module preserves the top-$V$\% most confident pseudo labels per category to maintain balanced supervision across all object classes, while the GALR module incorporates 3D geometric priors through superpoint analysis to improve label accuracy and boundary precision. 

Despite these refinements, significant portions remain unlabeled. Our second stage addresses this through Self-Training with Label Propagation (STLP), which iteratively trains the model using the refined pseudo labels. The STLP module combines a Label Update strategy and the GALR module to extend pseudo labels to unlabeled regions. Specifically, the Label Update strategy gradually propagates pseudo labels to unlabeled regions. It simultaneously retains the historical pseudo labels. In addition, it incorporates new predictions based on their reliability. These components are finally merged to generate the updated pseudo labels. The GALR module gradually improves the model by introducing 3D geometric priors to enhance the reliability of labels. During model inference, only the point cloud is simply input into the trained model to directly generate semantic segmentation predictions. Furthermore, the GALR module is employed to further refine these predictions, ultimately producing highly accurate semantic segmentation results. By integrating these components, our proposed approach establishes a more robust and geometry-aware framework for 3D WSSS. In summary, the main contributions of this paper are as follows:

\begin{itemize}
\item We propose a simple yet effective 3D weakly supervised semantic segmentation method that synergistically integrates 3D geometric priors and class-aware semantic cues to produce balancing and reliable point-level pseudo labels using only scene-level labels. 

\item  A Self-Training strategy is proposed to propagate pseudo labels to unlabeled regions by a integration of model self-training and 3D geometric priors to iteratively obtain high-quality pseudo labels, leading to a final robust 3D WSSS model.

\item Extensive experiments on the ScanNetv2 and S3DIS datasets demonstrate that our developed method achieves substantial performance improvements over previous state-of-the-art approaches. Notably, even when extended to unsupervised settings, our method maintains competitive performance, further validating its effectiveness and generalizability in leveraging geometric and semantic information for 3D scene understanding.
\end{itemize}

\section{Related Work}
\label{sec:work}
In this section, we provide a concise overview of existing research on vision-language models, 2D open-vocabulary semantic segmentation, 3D semantic segmentation, and self-training based methods.

\subsection{Vision-Language Models}
Exploring the interaction between vision and language is a fundamental research area in artificial intelligence. Vision-language models~\cite{kim2021vilt, radford2021learning, xu2023weakly, 10914541, liu2024less, jia2021scaling,10816351} seek to integrate textual semantics to improve performance across various vision tasks. Among them, Contrastive Language-Image Pretraining (CLIP)~\cite{radford2021learning} has gained prominence as a pivotal approach. CLIP employs dual encoders for images and text, trained through a contrastive learning paradigm to align visual and linguistic representations in a shared embedding space. During training, given a batch of image-text pairs, the model learns to associate each image with its corresponding textual description by maximizing their mutual similarity while reducing similarity with non-matching pair. Leveraging this robust alignment between 2D visual and textual modalities, CLIP achieves exceptional performance in zero-shot learning scenarios across a broad spectrum of vision tasks, underscoring its strong generalization capabilities and potential for effective transfer learning.

\subsection{2D Open-Vocabulary Semantic Segmentation}
Recent advancements in large-scale vision-language models have significantly enhanced the robustness and generalization capabilities of open-vocabulary semantic segmentation~\cite{liang2023open,xu2022simple,xu2023side,xu2023learning,chen2023exploring}. This challenging task focuses on segmenting target categories that remain unseen during training. Pioneering approaches like ZS3Net~\cite{bucher2019zero} utilize generative models to synthesize pixel-level features from word embeddings of novel classes, while SPNet~\cite{xian2019semantic} projects visual features into a shared semantic embedding space to align them with corresponding textual representations. More contemporary methods leverage the pretrained vision-language models such as CLIP~\cite{radford2021learning} to tackle open-vocabulary challenges. For instance, ZSSeg~\cite{xu2022simple} employs CLIP's visual encoder to generate class-agnostic segmentation masks and retrieves unseen class labels via its text encoder. OpenSeg~\cite{ghiasi2022scaling} further advances this paradigm by aligning segment-level visual features with text embeddings through region-word correspondence grounding.  In our work, we leverage pretrained 2D open-vocabulary models as the sole supervision source and extend their semantic understanding capabilities to 3D WSSS tasks. 

\subsection{3D Semantic Segmentation}
\subsubsection{\textbf{Fully Supervised Methods}}
Deep learning has catalyzed extraordinary advancements across diverse domains, particularly in image processing and computer vision. Recent breakthroughs have extended these capabilities to the demanding task of semantic segmentation in 3D point clouds, achieving remarkable effectiveness. A seminal work in this field is PointNet~\cite{qi2017pointnet}, which established the first neural architecture for point cloud semantic learning. PointNet employs shared multi-layer perceptrons (MLPs) to extract point-wise features and combines these with global features through aggregation, generating point-global representations for semantic prediction. However, due to its constrained capacity to capture local geometric structures, numerous point-based methods~\cite{qi2017pointnet++,10594720,9833393,9410334} have since emerged to enhance local feature representation. Additionally, voxel-based approaches~\cite{choy20194d,zhou2018voxelnet} segment point clouds into small voxels to better capture both local and global context, further facilitating semantic segmentation performance. Beyond these, recent methods~\cite{sun2023superpoint, li2025cross, yin2024sai3d} introduce additional geometric priors into the learning process. For instance, certain methods utilize normal-based graph cut algorithms~\cite{landrieu2018large} to over-segment point clouds and extract boundary information, which serves as a prior to guide networks in learning geometry-aware features. In this paper, we propose GALR, a novel approach that utilizes superpoints as auxiliary geometric cues to assist the model in learning meaningful geometric priors. Distinct from previous approaches, our method operates under a substantially weaker supervision paradigm—requiring only scene-level annotations. Moreover, rather than depending on feature distances, we also design a geometric voting mechanism with a majority-class constraint, which together produce more reliable and higher-quality pseudo labels for 3D semantic segmentation, advancing the state-of-the-art in weakly supervised point cloud analysis. 

\subsubsection{\textbf{Weakly Supervised Methods}}
Recent advances in 3D WSSS for point clouds focus on reducing annotation costs through weaker supervision signals, including sparsely labeled points~\cite{hu2022sqn,zhang2022not}, box-level annotations~\cite{chibane2022box2mask}, subcloud-level labels~\cite{wei2020multi}, and scene-level supervision~\cite{ren20213d,kweon2022joint,yang20232d,xu20243d}. Among these, scene-level annotations have gained particular attention due to their minimal annotation requirements. WyPR~\cite{ren20213d} first demonstrates the feasibility of learning 3D semantic segmentation using only scene-level labels. Kweon \textit{et al.}~\cite{kweon2022joint} further incorporates 2D RGB images with corresponding image-level labels to guide the 3D WSSS model. However, the additional cost of image-level annotations motivated MIT~\cite{yang20232d} to develop a transformer-based approach that implicitly aligns 2D and 3D embeddings without geometric camera calibration. Above methods predominantly rely on class activation map solutions for 3D WSSS, but these face significant challenges due to the large-scale of 3D scenes, often leading to imprecise activation regions and underutilized category-specific information. To address this, Xu \textit{et al.}~\cite{xu20243d} introduced 3DSS-VLG, which leverages pretrained vision-language models for 3D training guidance. Nevertheless, 3DSS-VLG overlooks intrinsic 3D point cloud priors that are particularly valuable for semantic segmentation. In contrast, our proposed method integrates geometric prior knowledge with self-training mechanisms to enhance 3D WSSS performance under weak supervision constraints.

\subsection{Self-Training based Methods}
Self-training has emerged as a prominent semi-supervised learning paradigm that utilizes pseudo labels generated on unlabeled data to iteratively enhance model performance. By propagating a small set of initial annotations to extensive unlabeled regions, this strategy has demonstrated remarkable efficacy across diverse domains, including 2D image understanding~\cite{9531449,10128961,10418852, melas2021pixmatch}, natural language processing~\cite{he2019revisiting}, and 3D scene comprehension~\cite{xiao2024domain,saltori2022cosmix,yang2021st3d}. A critical challenge in self-training pertains to designing effective mechanisms for updating pseudo labels and reliably propagating predictions to unlabeled areas. Recent advancements have introduced innovative solutions to address these challenges. For instance, Melas-Kyriazi \textit{et al.}~\cite{melas2021pixmatch} incorporate consistency regularization to maintain pseudo label stability under input perturbations. Xie \textit{et al.}~\cite{xie2022simmim} enhance feature representations through contrastive learning-based self-supervised pretraining. Other approaches~\cite{caron2021emerging,tarvainen1703weight} adopt teacher-student frameworks, where the teacher model serves as an exponential moving average of the student, improving resilience to noisy pseudo labels. In this paper, we propose a self-training framework that incorporates a Label Update strategy with the GALR module to progressively propagate and refine pseudo labels across unlabeled 3D spaces, leading to improved segmentation performance under weak supervision. 

\section{The Proposed Methodology}
\begin{figure*}
\centering
\includegraphics[width=1.0\linewidth]{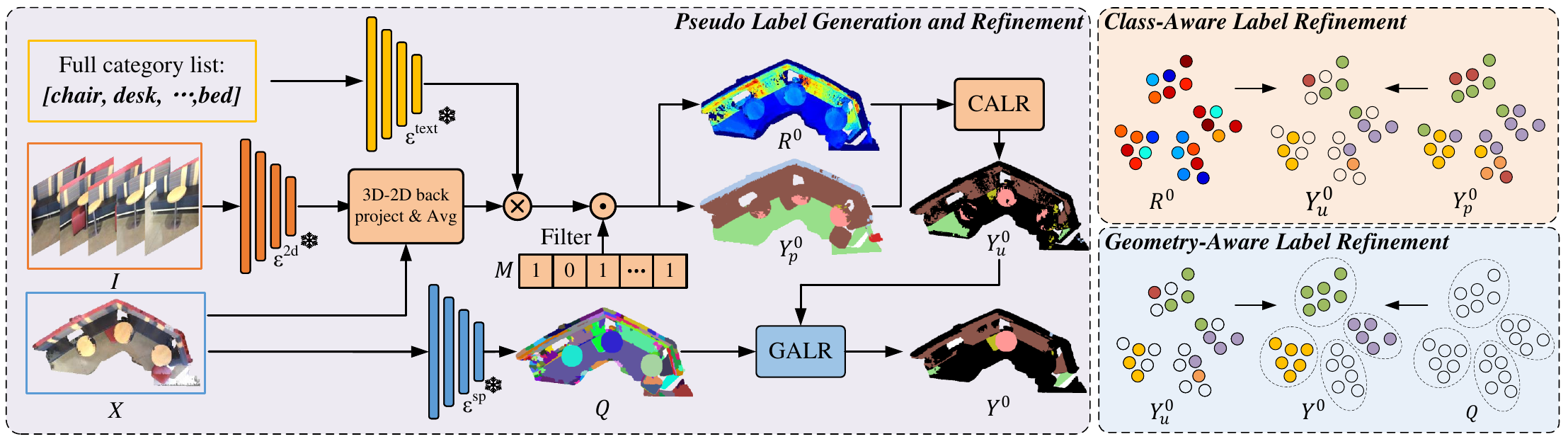} 
\caption{\small The proposed Pseudo Label Generation and Refinement procedures. We first extract 2D embeddings $F_{2D}$ and text embeddings $F_C$ using a pretrained VLM. The 2D embeddings are back-projected via camera calibration to obtain 2D-projected embeddings $P_{2D}$. Prediction logits are computed by multiplying $F_C$ and $P_{2D}$, then filtered with the scene-level label mask $M$. After ranking, initial pseudo labels ${Y^0_p}$ and confidence scores ${R^0}$ are obtained. CALR selects top-$V$\% pseudo labels per category based on $R^0$ to ensure class balance and confidence. GALR refines labels by superpoint overlap: if a dominant category in a superpoint exceeds a threshold, the block is assigned that category; otherwise, it remains unlabeled. The final pseudo labels $Y^0$ are obtained after refinement. Notably, in $R^0$, color depth represents confidence, akin to a heatmap, where darker colors indicate higher confidence.  Points with the same color in the pseudo labels $Y^0$ correspond to the same predicted category, and the dotted circle in $Q$ denotes points within the same superpoint.} 
\label{fig2}
\end{figure*}

In this paper, we devise a new weakly supervised method for 3D semantic segmentation, comprising two core components: Pseudo Label Generation and Refinement procedure and Self-Training with Label Propagation. As illustrated in Fig.~\ref{fig2}, the Pseudo Label Generation and Refinement procedure is utilized to produce high-fidelity point-level pseudo labels under 3D WSSS with only scene-level supervision. Subsequently, as shown in Fig.~\ref{fig5}, the STLP component propagates these refined pseudo labels to unlabeled regions and iteratively optimizes the model through self-training cycles using the progressively refined labels.

\subsection{Pseudo Label Generation}
\label{ppg}
Following ~\cite{xu20243d,peng2023openscene}, we use a pretrained VLM~\cite{ghiasi2022scaling,lilanguage} and scene-level labels to generate pseudo labels with associated probabilities, as shown in  Fig.~\ref{fig2}. The input consists of a 3D point cloud, multi-view images, and scene-level labels. The point cloud scene, ${X} \in \mathbb{R}^{N \times 6}$, contains $N$ points, each represented by six dimensions (RGBXYZ). The multi-view RGB images, $I$, consist of $L$ images with a resolution of $H \times W$. The scene-level label mask $M\in \mathbb{R}^{K}$, where $K$ denoted the number of categories.

First, we apply the image encoder of a pretrained vision-language model~\cite{ghiasi2022scaling,lilanguage} to extract per-pixel 2D embeddings, denoted as ${F}_{2D} \in \mathbb{R}^{L \times H \times W \times d}$, where $d$ is the 2D embedding dimension. For each point in the 3D point cloud, we compute its corresponding 2D position using the intrinsic and extrinsic matrices. We then extract the projected 2D embeddings from ${F}_{2D}$ based on these calculated 2D positions. Since a point may have multiple correspondences across different images, the final 2D-projected embeddings, ${P}_{2D} \in \mathbb{R}^{N \times d}$, are obtained by averaging all corresponding 2D embeddings. Specifically, given the $n$-\textit{th} point $(x_{3D}^n,y_{3D}^n,z_{3D}^n) \in \mathbb{R}^{3}$ in the point cloud, we project it onto the $i$-\textit{th} image $I_i\in\mathbb{R}^{H \times W \times 3}$. The projection position $(x_{2D},y_{2D}) \in \mathbb{R}^{2}$ on the image can be computed as:

\begin{equation}
z\cdot\begin{bmatrix}x_{2D}\\y_{2D}\\1\end{bmatrix} = CK\cdot\begin{bmatrix}CR&CT\\0&1\end{bmatrix} \cdot \begin{bmatrix}x_{3D}\\y_{3D}\\z_{3D}\\1\end{bmatrix},
\end{equation}
where $CK$ represents the camera intrinsic matrix, while the rotation matrix $CR$ and the translation vector $CT$ define the camera extrinsic parameters.  

Subsequently, for the corresponding 2D embedding $F_{2D}^i\in\mathbb{R}^{H \times W \times d}$, if the projected point falls within the image grid, we extract the corresponding projected embedding ${f}_{2D}^{ij} \in\mathbb{R}^{1 \times d}$ from ${F}_{2D}^i$. Since each point may have multiple correspondences across different images, the final 2D-projected embedding for a point ${p}_{2D}^n\in\mathbb{R}^{1\times d}$ is obtained by averaging all its associated embeddings:

\begin{equation}
    p_{2D}^n = \sum_{j=0}^{J} f^{ij}_{2D}.
\end{equation}
With regard to a point cloud ${X}$, we process each point following the steps above, obtaining the 2D-projected embeddings $P_{2D} = \left \{ p_{2D}^1, p_{2D}^2, \dots, p_{2D}^N \right \} \in\mathbb{R}^{N\times d}$.  

Moreover, we use the text encoder of the pretrained model to extract the text embeddings ${F}_{C} \in \mathbb{R}^{C \times d}$ for all category labels, where $C$ is the number of categories. We then compute the classification logits, ${L}_{2D} \in \mathbb{R}^{N \times C}$, by performing matrix multiplication between the text embeddings ${F}_{C}$ and the 2D-projected embeddings ${P}_{2D}$. To refine these logits, we compute the inner product between ${L}_{2D}$ and the scene-level label mask ${M}$, yielding the filtered logits ${L}_f \in \mathbb{R}^{N \times C}$, where ${M} \in \mathbb{R}^{1 \times C}$ is a boolean mask indicating valid scene categories. Finally, after ranking the ${L}_f$, we generate the pseudo labels ${Y^0_p} \in \mathbb{R}^{N}$ and their corresponding probabilities ${R^0} \in \mathbb{R}^{N}$.  

\subsection{Pesudo Label Refinement}
\label{plr}

\begin{figure}[t]
\centering
\includegraphics[width=1.0\linewidth]{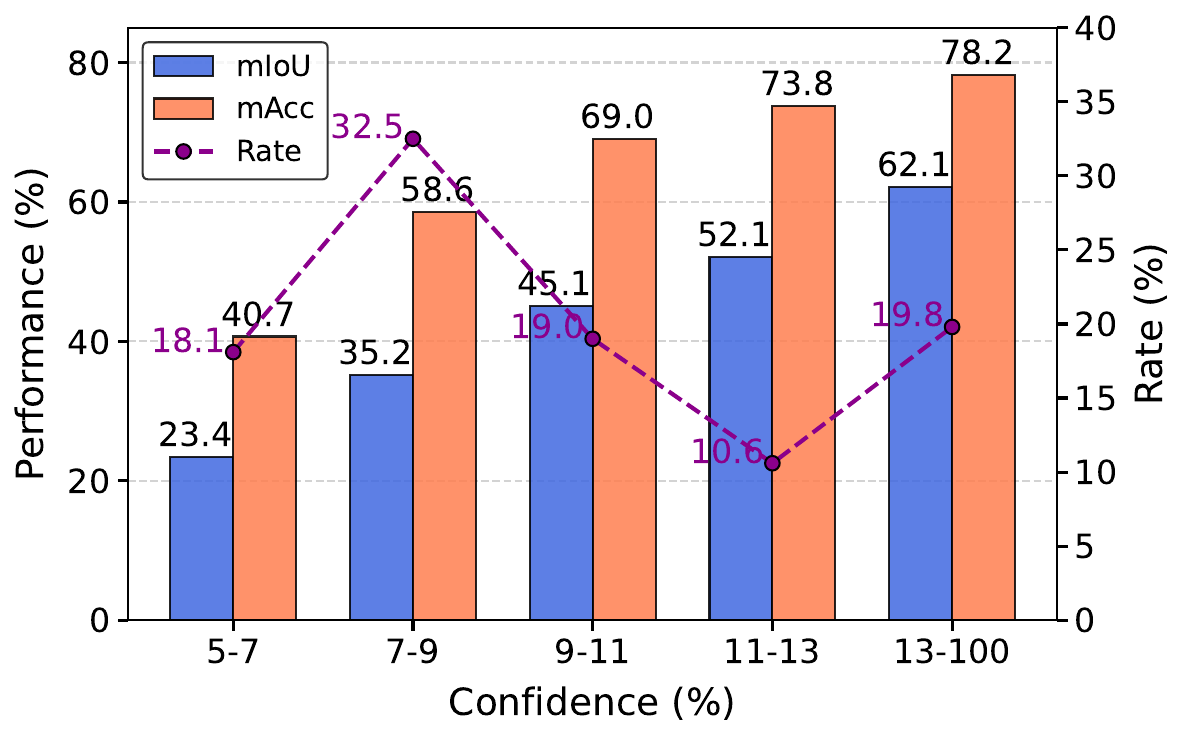}
\caption{\small We present the performance of pseudo labels across different confidence intervals, along with the proportion of all points within each changed confidence range. The results indicate that higher confidence levels correspond to better segmentation performance. Notably, more than half of the pseudo labels exhibit confidence below 9\%, highlighting the ambiguity of the original pseudo labels.} 
\label{fig3}
\end{figure}

Although the filtering strategy can effectively enhance the initial pseudo labels, there remain some limitations. On the one hand, as shown in  Fig.~\ref{fig3}, compared to high-confidence pseudo labels, low-confidence pseudo labels are more likely to be inaccurate. On the other hand, the current approach relies heavily on the pretrained VLM, neglecting inherent 3D geometric priors. To address these challenges, we introduce Class-Aware Label Refinement and Geometry-Aware Label Refinement, two synergistic strategies that systematically integrate class-aware semantic context with 3D geometric priors for robust label optimization.

\subsubsection{\textbf{Class-Aware Label Refinement}} 
\begin{figure}[t]
  \centering
   \includegraphics[width=1\linewidth]{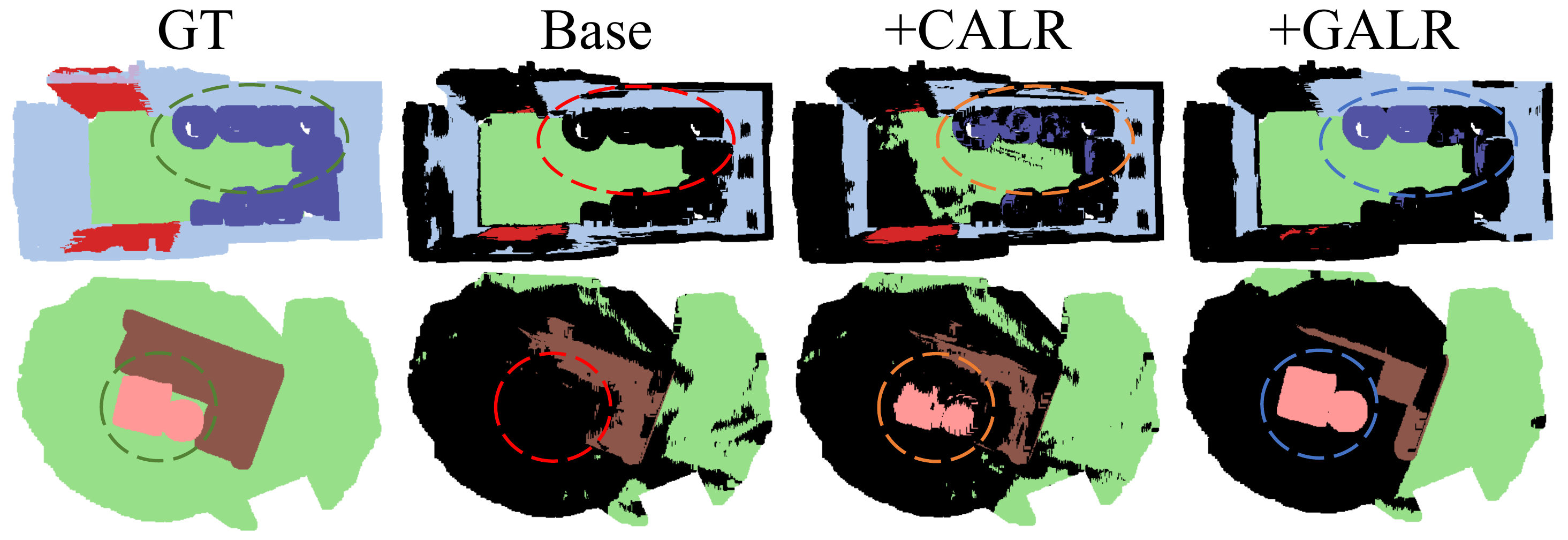}
   \caption{\small Visualization of applying global of top-$V$\% selection (Base), our CALR and GALR on ScanNet dataset. From left to right: ground truth, global top-$V$\%, our CALR results and GALR results.}
   \label{fig4}
\end{figure}

Low-confidence predictions in pseudo labels ${Y^0_p}$ are more prone to inaccurate. A straightforward approach might be to retain the top-$V$\% of points based on confidence. However, this exacerbates the class imbalance problem, as larger categories (\textit{e.g.}, floors, walls) dominate, leaving smaller categories underrepresented, as depicted in Fig.~\ref{fig4}. Imbalanced pseudo labels can negatively impact model training, leading to a loss of segmentation capability for small-category objects.% as shown in Tab.~\ref{Tab.5}.

Motivated by ~\cite{zou2018unsupervised, he2021re}, we develop the CARL strategy. Rather than applying global top-$V$\% selection, we perform the selection within each class. This ensures that high-confidence points from both large and small categories are retained, preventing the over-representation of dominant categories. After selecting the top-$V$\% points for each class, the remaining low-quality and low-confidence points are set to an unlabeled state $\beta$, yielding the refined labels ${Y_u^0}$. This process enhances label accuracy and ensures a more balanced distribution across categories.

\begin{algorithm}[t]
\caption{Geometry-Aware Label Refinement}
\label{alg:sup_point_ensemble}
\renewcommand{\algorithmicrequire}{\textbf{Input:}}
\renewcommand{\algorithmicensure}{\textbf{Output:}}
\begin{algorithmic}[1]
    \REQUIRE Initial pseudo labels ${Y_u^T} \in \mathbb{R}^N$; \\ \quad \  Superpoints $\left\{Q_i\right\}_{i=1}^{U}$; \\ \quad \ Overlap threshold $\alpha \in [0,1]$; \\ \quad \  Unlabeled tag $\beta$;
    \ENSURE Refined pseudo labels ${Y^{T}}$
    
    \STATE Initialize output pseudo labels $\left\{Y_i^{T}\right\}_{i=1}^{U} \in \{\beta\}^{N}$
    
    \FOR{ $i = 1$ to $U$}   
            \STATE Initialize counter ${A} \in \{0\}^{C}$
            \STATE Initialize overlap pseudo labels $O \gets Q_i \cap Y_u^T$
            \STATE $A \gets \textbf{Count} (O)$ 

            \STATE $r \gets \textbf{max}(A) / \textbf{sum}(A)$  
            
            \IF {$r > \alpha$}
                \STATE ${Y_i^{T}} \gets \textbf{argmax}(A)$
            \ELSE
                \STATE ${Y_i^{T}} \gets \beta $
            \ENDIF
    \ENDFOR
\RETURN ${Y^{T}}$
\end{algorithmic}
\end{algorithm}

\begin{figure*}[t]
\centering
\includegraphics[width=1.0\linewidth]{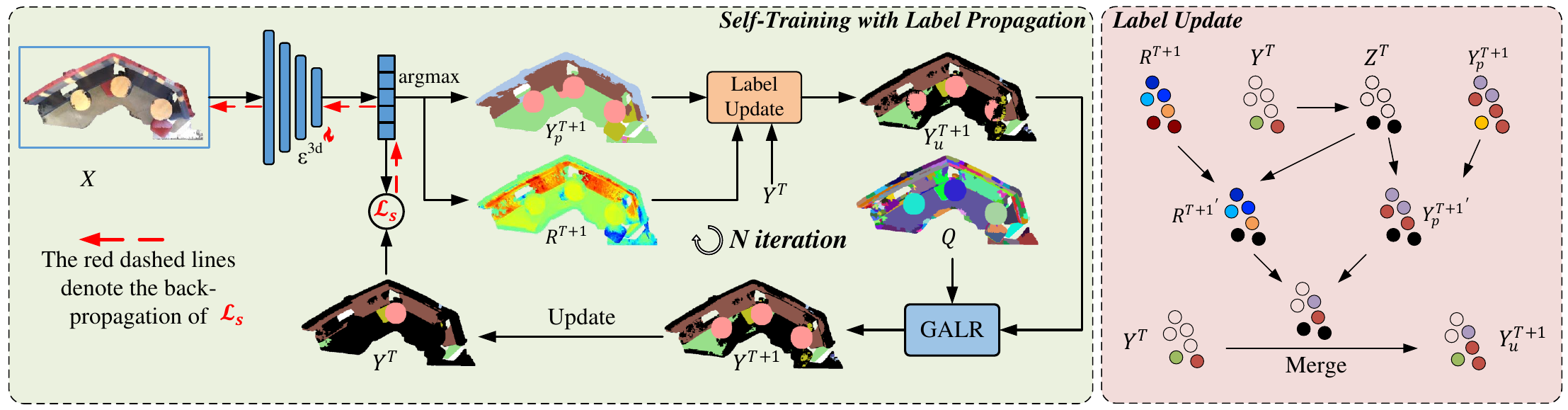} % Reduce the figure size so that it is slightly narrower than the column.
\small
\caption{\small The proposed Self-Training with Label Propagation follows an iterative approach. First, the model is trained using pseudo labels $Y^T$ from the previous step. Then, inference on the training set generates updated predictions $Y^{T+1}_p$ and confidence scores $R^{T+1}$. To propagate pseudo labels to unlabeled regions, the Label Update strategy retains previous pseudo labels  $Y^T$ while incorporating reliable new predictions to generate updated pseudo labels $Y^{T+1}_u$. GALR further refines them to obtain the final updated pseudo labels $Y^{T+1}$. This process iterates, using updated pseudo labels as supervision, progressively extending labels to unlabeled regions. Notably, in $R^{T+1}$, color depth represents confidence levels.  Points with the same color in the pseudo labels $Y^T$ and $Y_p^{T+1}$ correspond to the same predicted category, while black-colored points denote masked regions that do not require any updates.
} 
\label{fig5}
\end{figure*}

\subsubsection{\textbf{Geometry-Aware Label Refinement}} Previous method relies solely on the pretrained VLM, overlooking the 3D geometric priors. To further improve the quality of the pseudo labels, we introduce a GALR strategy that incorporates 3D geometric priors. The detailed procedure is presented in Algorithm~\ref{alg:sup_point_ensemble}. Specifically, following~\cite{sun2023superpoint, yin2024sai3d}, we apply a normal-base graph cut algorithm $\varepsilon^{sp}$\cite{landrieu2018large} to over-segment the point cloud $X$ in to a set of superpoint $\left\{Q_i\right\}_{i=1}^{U}$, which group nearby points sharing similar geometric features. For each superpoint block $Q_i$, we calculate the overlap between the pseudo labels $Y_u^0$ and it to get the the intersection pseudo labels $O$. Then, we compute the category distribution matrix $A$ in $O$ and obtain the rate $r$ of the most frequent category. If it surpasses the overlap threshold $\alpha$, we will assign the most frequent category of intersection pseudo labels $O$ to the output pseudo labels $Y^0_i$. Otherwise it demonstrates no category dominates the block, indicating ambiguity in the most frequent category, the pseudo labels $Y^0_i$ will be set to an unlabeled state $\beta$. This process ensures that the pseudo labels are consistent with the majority of points in the block, while avoiding incorrect label assignments in ambiguous cases. 

Through the CALR and GALR strategies, the final pseudo labels ${Y^0}$ are refined by integrating class-aware information from ${Y_u^0}$ with 3D geometric priors, resulting in more accurate and reliable labels for 3D semantic segmentation.

\subsection{Self-Training with Label Propagation}
\label{st}
Although the accuracy of the refined pseudo labels, ${Y^0}$, is sufficiently high for labeled points, there remain large areas of unlabeled points. To facilitate network training, we propose the Label Update strategy and leverage the self-training strategy to propagate labels to unlabeled regions. Concretely, as illustrated in Fig.~\ref{fig5}, we first train the 3D module $\varepsilon^{3d}$ with the pseudo labels ${Y^{T}}$ of previous step. The point cloud ${X}$ is assigned as input, and MinkowskiNet18A UNet~\cite{choy20194d} is utilized as the 3D module to obtain the point-level classification logits $L_{3D}$. Subsequently, we utilize the pseudo labels ${Y^{T}}$ as supervisory and introduce the cross-entropy loss $\mathcal{L}_s$ to supervise the model. After the training stage, we perform inference on the training data set to obtain the predicted labels ${Y^{T+1}_p}$ and the probabilities of points ${R^{T+1}}$. Here, we utilize the scene-level mask $M$ to filter the logits.

Secondly, we update the label of the previous step, ${Y^{T}}$ into ${Y^{T+1}}$ via Label Propagation procedure, which consists with Label Update strategy and GALR strategy. Specifically, as shown in Algorithm~\ref{alg2}, in the Label Update stage, we first utilize the scene-level label mask $M$ to filter the pseudo labels ${Y^{T+1}_p}$ and point probabilities ${R^{T+1}}$. Then we retain the previous pseudo labels ${Y^{T}}$ and generate the mask ${Z^{T}}$ to indicate which points need updating. The matrix inner product of ${Y^{T+1}_p}$ and ${R^{T+1}}$ with ${Z^{T}}$ is performed to get the masked ${Y^{{T+1}^{'}}_p}$ and ${R^{{T+1}^{'}}}$. Besides, following the CALR strategy, we also retain the top-$V$\% highest-ranked probabilities within each category in ${R^{{T+1}^{'}}}$.  After the new retrained pseudo labels are obtained, they are merged with the previous pseudo labels  ${Y^{T}}$ to form the updated pseudo labels ${Y^{T+1}_u}$. The GALR strategy is to further employed to refine ${Y^{T+1}_u}$. Subsequently, the final updated pseudo labels ${Y^{T+1}}$ are generated, where previously unlabeled regions are now equipped with reliable pseudo labels.

Notably, our proposed STLP component operates iteratively, with each cycle refining pseudo-label precision, propagating labels to previously unlabeled regions, and strengthening the training process to achieve superior point cloud segmentation performance.

\subsection{Inference}
\label{inf}
During inference, our method operates exclusively on 3D point clouds, requiring no auxiliary 2D images. We utilize the final trained model from the STLP procedure and input the point cloud into the model to obtain the predicted segmentation.
% The final model trained through the Self-Training with Label Propagation framework processes the raw point cloud to generate initial segmentation predictions. This model has been iteratively refined during STLP training to capture robust geometric features. 
Subsequently, the GALR  strategy is applied to enhance prediction coherence by leveraging spatial relationships and geometric constraints. This post-processing step resolves local ambiguities and sharpens semantic boundaries. This combined pipeline achieves efficient 3D segmentation, producing accurate per-point labels.

\begin{algorithm}[t]
\caption{Label Update}
\label{alg:label_propagation}
\renewcommand{\algorithmicrequire}{\textbf{Input:}}
\renewcommand{\algorithmicensure}{\textbf{Output:}}
\begin{algorithmic}[1]
    \REQUIRE Previous pseudo labels $Y^T \in \{1, \dots, C\}^N$; \\ \quad \  Predicted labels $Y_p^{T+1} \in \{1, \dots, C\}^N$; \\ \
    \quad Predicted class probabilities $R^{T+1} \in [0,1]^{N \times C}$; \\ \
    \quad Scene-level mask $M \in \{0,1\}^N$; \\ \
    \quad Update mask $Z^T \in \{0,1\}^N$; \\ 

    \ENSURE Updated pseudo labels $Y_u^{T+1} \in \{1, \dots, C\}^N$
    
    \STATE $R^{T+1} \leftarrow R^{T+1} \cdot M$, \quad $Y_p^{T+1} \leftarrow Y_p^{T+1} \cdot M$

    \STATE $Z_i^T \leftarrow 1$ if $Y^T_i = \beta$, else $0$
    \STATE $Y^{{T+1}^{'}}_p \leftarrow Z^T \cdot Y_p^{T+1}$, \quad ${R^{{T+1}^{'}}} \leftarrow Z^T \cdot R^{T+1}$
    \STATE $Y^{{T+1}^{'}}_p \leftarrow CALR(Y^{{T+1}^{'}}_p, {R^{{T+1}^{'}}})$
    \STATE $Y_{u,i}^{T+1} \leftarrow 
        \begin{cases}
        Y^{{T+1}^{'}}_p, & \text{if } Z_i^T = 1 \\
        Y_i^T, & \text{otherwise}
        \end{cases}$

    \RETURN $Y_u^{T+1}$
\end{algorithmic}
\label{alg2}
\end{algorithm}

\section{Experiments}
\subsection{Experimental Settings}
\subsubsection{Datasets and Evaluation Metrics}
We evaluate our proposed method on two widely-used benchmarks, ScanNet~\cite{dai2017scannet} and S3DIS~\cite{armeni20163d}. The ScanNet dataset comprises 1513 training scenes and 100 test scenes across 20 semantic categories. Following the official train-val split, we utilize 1,205 scenes for training and 312 scenes for validation. S3DIS contains 6 areas with 271 rooms, each captured by RGBD sensors and represented as 3D point clouds with XYZ coordinates and RGB attributes. Consistent with prior works, we adopt Area 5 for testing. Performance is measured using the mean Intersection-over-Union (mIoU) metric across all categories, which quantifies the overlap between predicted labels and ground truth labels.

\subsubsection{Implementations Details}
In the experiment, the hyperparameters $V$ of the CALR strategy is set to 30. And the overlap threshold $\alpha$ in the GALR strategy is set to 0.5. In addition, during the STLP procedure, we use the SGD optimizer with a base batch size of 4 and initialize the learning rate to 0.01. The learning rate is adjusted using the poly learning rate policy, and the hyperparameter $T$ is set to 2. Our method is implemented using PyTorch.

\subsection{3D Semantic Segmentation Results}

\begin{table}[t]
\caption{\small Performance comparison on the ScanNet val set and test set. ``Sup.'' indicates the type of supervision. ``100\%'' represents full annotation. ``subcloud.'' and ``scene.'' imply subcloud-level annotation and scene-level annotation respectively. ``image.'' denotes image-level annotation. $^{\dagger}$ indicates results reproduced by us.}
\begin{center}
\small
\setlength{\tabcolsep}{0.35mm}{
\begin{tabular}{ccccc}
\toprule %绘制一条水平横线
Method                                         & Label Effort                                & Sup.     & Val   & Test  \\
\hline %绘制一条水平横线
PointNet++~\cite{qi2017pointnet++}             & \multirow{5}{*}{\textgreater 20 min}        & 100\%    & -     & 33.9  \\
MinkowskiNet~\cite{choy20194d}                 &                                             & 100\%    & 72.2  & 73.6  \\
KPConv~\cite{thomas2019kpconv}                 &                                             & 100\%    & 69.2  & 68.6  \\
PointNetXt~\cite{qian2022pointnext}            &                                             & 100\%    & 71.5  & 71.2  \\
DeepViewAgg~\cite{robert2022learning}          &                                             & 100\%    & 71.0  & -     \\
\midrule%绘制一条水平横线
MPRM~\cite{wei2020multi}                       & 3 min                                      & subcloud. & 43.2  & 41.1  \\
\midrule %绘制一条水平横线
Kweon \textit{et al}.~\cite{kweon2022joint}    & 5 min                            & scene. + image.     & 49.6   & 47.4  \\
\midrule %绘制一条水平横线
MIL-Trans~\cite{yang2022mil}                   &  \multirow{7}{*}{\textless 1 min}          & scene.    & 26.2   & -     \\
WYPR~\cite{ren20213d}                          &                                            & scene.    & 29.6   & 24.0  \\
MIT~\cite{yang20232d}                          &                                            & scene.    & 35.8   & 31.7  \\
3DSS-VLG(OpenSeg)~\cite{xu20243d}              &                                            & scene.    & 49.7   & 48.9  \\
3DSS-VLG(LSeg)$\dagger$~\cite{xu20243d}                 &                                            & scene.    & 55.4   & 53.8  \\
Ours(Openseg)                                  &                                            & scene.    & 57.5   & 56.6  \\
Ours(Lseg)                                     &                                            & scene.    & \textbf{64.1} & \textbf{62.5}  \\
\bottomrule %绘制一条水平横线
\end{tabular}}
\label{Tab.1}
\end{center}
\end{table}

\subsubsection{Evaluation on ScanNet}
Table~\ref{Tab.1} presents a performance comparison of the 3D point cloud semantic segmentation methods evaluated on the ScanNet dataset. Compared to scene-level annotation-supervised approaches, we can find that our proposed method significant superiority over the current state-of-the-art method 3DSS-VLG~\cite{xu20243d}. Furthermore, when evaluated against other weakly supervised methods that utilize richer supervision signals (such as subcloud-level annotations or additional image-level annotations), our approach achieves remarkable improvements: outperforming MPRM ~\cite{wei2020multi} by 20.9\% and 21.4\%, and surpassing Kweon \textit{et al.}'s method~\cite{kweon2022joint} by 14.5\% and 15.1\% on the validation and test datasets respectively. We also provide class-wise segmentation performance comparisons in Table~\ref{Tab.3}. From Table~\ref{Tab.3}, it is obvious that our proposed method obtains better performance than other methods.These findings demonstrate that enhancing pseudo-label quality and leveraging previously underutilized 3D geometric priors can substantially improve model performance.

Additionally, we analyze the impact of different VLMs in Table~\ref{Tab.1}, specifically comparing OpenSeg~\cite{ghiasi2022scaling} and LSeg~\cite{lilanguage}. While model performance varies with the choice of pre-trained model, our proposed  method maintains consistent effectiveness across different VLMs. Besides, we also compare our method with several fully supervised methods. On the one hand, our method achieves even superior results over the fully supervised methods~\cite{qi2017pointnet}. On the other hand, compared with the time consumption of supervised signal annotation, we can find that our scene-level annotation cost is much less than the full supervision annotation cost. Compared to MinkowskiNet~\cite{choy20194d}, which shares the same architecture but uses full supervision, our method achieves only 8.1\% lower performance on the validation set. This underscores the effectiveness and promising potential of our weakly supervised approach, particularly in balancing performance with annotation cost efficiency.

\begin{table}[t]
\caption{\small Performance comparison on the S3DIS dataset. ``Sup.'' indicates the type of supervision. ``100\%'' represents full annotation. ``scene.'' denotes scene-level annotation.}
\begin{center}
\small
\begin{tabular}{cccc}
\toprule %绘制一条水平横线
Method                                         & Label Effort                         & Sup.                                  & Test \\
\hline %绘制一条水平横线
PointNet~\cite{qi2017pointnet}                 & \multirow{7}{*}{\textgreater 20 min} & 100\%                                 & 41.1 \\
TangentConv~\cite{tatarchenko2018tangent}      &                                      & 100\%                                 & 52.8 \\
MinkowskiNet~\cite{choy20194d}                 &                                      & 100\%                                 & 65.8 \\
KPConv~\cite{thomas2019kpconv}                 &                                      & 100\%                                 & 67.1 \\
PointTransformer~\cite{zhao2021point}          &                                      & 100\%                                 & 70.4 \\
PointNetXt~\cite{qian2022pointnext}            &                                      & 100\%                                 & 70.5 \\
DeepViewAgg~\cite{robert2022learning}          &                                      & 100\%                                 & 67.2 \\
\midrule %绘制一条水平横线
MPRM~\cite{wei2020multi}                       & \multirow{6}{*}{\textless 1 min}     & scene.                                & 10.3 \\
MIL-Trans~\cite{yang2022mil}                   &                                      & scene.                                & 12.9 \\
WYPR~\cite{ren20213d}                          &                                      & scene.                                & 22.3 \\
MIT~\cite{yang20232d}                          &                                      & scene.                                & 27.7 \\
3DSS-VLG~\cite{xu20243d}                       &                                      & scene.                                & 45.3 \\
Ours                                           &                                      & scene.                                & \textbf{51.8}\\
\bottomrule %绘制一条水平横线
\end{tabular}
\label{Tab.2}
\end{center}
\end{table}

\begin{table*}[htbp]
\caption{Class-wise IoU on ScanNet validation set. For simplicity, we abbreviate cabinet/window/bookshelf/picture/counter/curtain/shower curtain/other furniture as cab./win./B.S./pic./cnt./cur./S.C./O.F., respectively. The OS and LS indicates the OpenSeg and LSeg, respectively.}
\begin{center}
\setlength{\tabcolsep}{1.0mm}{\begin{tabular}{c|cccccccccccccccccccc|c}
 \toprule
Method & wall & floor & cab. & bed & chair & sofa & table & door & win. & B.S. & pic. & cnt. & desk & cur. & fridge & S.C. & toilet & sink & tub & O.F & mIoU \\
 \midrule
WyPR~\cite{ren20213d} & 58.1 & 33.9 & 5.6 & 56.6 & 29.1 & 45.5 & 19.3 & 15.2 & 34.2 & 33.7 & 6.8 & 33.3 & 22.1 & 65.6 & 6.6 & 36.3 & 18.6 & 24.5 & 39.8 & 6.6 & 29.6 \\
MPRM~\cite{wei2020multi} & 59.4 & 59.6 & 25.1 & 64.1 & 55.7 & 58.7 & 45.6 & 36.4 & 40.3 & 67.0 & 16.1 & 22.6 & 42.9 & \textbf{66.9} & 24.1 & 39.6 & 47.0 & 21.2 & 44.7 & 28.0 & 43.2 \\
Kweon \textit{et al.}~\cite{kweon2022joint} & 69.6 & 90.0 & 27.9 & 61.0 & 68.7 & 62.7 & 52.3 & 34.1 & 42.0 & 65.2 & 5.8 & 42.6 & 44.4 & 60.4 & 25.3 & 33.5 & 70.9 & 38.6 & 66.5 & 31.4 & 49.6 \\
3DSS-VLG (OS)~\cite{xu20243d} & 67.6 & 82.8 & 44.6 & 68.0 & 63.0 & 58.7 & 43.6 & 42.5 & 44.4 & 67.5 & 18.0 & 22.6 & 32.8 & 63.0 & 40.0 & 33.9 & 76.1 & 33.0 & 69.8 & 23.0 & 49.7 \\
3DSS-VLG (LS) & 73.5 & 89.1 & 46.3 & 73.4 & 69.6 & 71.6 & 47.7 & 47.0 & 49.6 & 62.2 & 18.4 & 52.1 & 41.8 & 63.5 & 41.9 & 52.4 & 89.7 & 17.2 & 83.6 & 18.1 & 55.4 \\
 \midrule
Ours (OS) & 75.6 & 92.0 & 48.4 & 75.0 & 71.6 & 73.2 & 49.0 & 48.4 & 50.7 & 63.2 & 20.6 & \textbf{56.6} & 44.2 & 64.5 & 43.8 & 54.1 & \textbf{93.9} & 18.2 & 88.0 & 18.7 & 57.5 \\
Ours (LS) & \textbf{78.8} & \textbf{96.0} & \textbf{52.2} & \textbf{78.8} & \textbf{84.4} & \textbf{80.6} & \textbf{66.4} & \textbf{54.2} & \textbf{54.7} & \textbf{73.9} & \textbf{25.5} & 55.5 & \textbf{49.7} & 63.6 & \textbf{44.2} & \textbf{60.8} & 90.0 & \textbf{54.7} & \textbf{89.6} & \textbf{27.7} & \textbf{64.1} \\
\bottomrule
\end{tabular}}
\label{Tab.3}
\end{center}
\end{table*}

\begin{table*}[!t]
\caption{The impact of pseudo-labels category imbalance. Here we provide class-wise IoU about pseudo labels and predictions on ScanNet dataset. For simplicity, we abbreviate cabinet/window/bookshelf/picture/counter/curtain/shower curtain/other furniture as cab./win./B.S./pic./cnt./cur./S.C./O.F., respectively.}
\begin{center}
\setlength{\tabcolsep}{1.15mm}{
\begin{tabular}{c|c|cccccccccccccccccccc|c}
\toprule
Method & Split & wall & floor & cab. & bed & chair & sofa & table & door & win. & B.S. & pic. & cnt. & desk & cur. & fridge & S.C. & toilet & sink & tub & O.F & mIoU \\
 \midrule
(a) & Train & \textbf{90.6} & \textbf{95.1} & \textbf{75.8} & \textbf{92.5} & 82.9 & 89.9 & 60.7 & 70.5 & 57.0 & 80.4 & 4.6 & 42.2 & 52.3 & \textbf{91.3} & \textbf{89.9} & 0 & 93.2 & \textbf{78.5} & \textbf{94.6} & 0.2 & 67.1 \\
(b) & Train & 88.8 & 93.8 & 70.5 & 92.3 & \textbf{83.5} & \textbf{91.0} & \textbf{74.1} & \textbf{72.3} & \textbf{72.1} & \textbf{85.1} & \textbf{62.3} & \textbf{71.0} & \textbf{69.0} & 87.9 & 84.1 & \textbf{74.4} & \textbf{93.2} & 73.9 & 91.9 & \textbf{21.4} & \textbf{77.6} \\
 \midrule
(a) & Val & 71.1 & 90.3 & 44.0 & 62.8 & 72.3 & 63.9 & 57.0 & 46.2 & 44.4 & 62.2 & 0.9 & 38.3 & 38.6 & 59.3 & 39.4 & 0 & 70.1 & 48.6 & 79.5 & 0 & 49.4 \\
(b) & Val & \textbf{77.6} & \textbf{93.9} & \textbf{46.8} & \textbf{72.8} & \textbf{78.2} & \textbf{76.0} & \textbf{62.1} & \textbf{51.9} & \textbf{52.3} & \textbf{72.6} & \textbf{18.5} & \textbf{53.8} & \textbf{45.2} & \textbf{64.5} & \textbf{41.3} & \textbf{51.4} & \textbf{76.8} & \textbf{54.7} & \textbf{85.9} & \textbf{23.4} & \textbf{60.0} \\
\bottomrule
\end{tabular}}
\label{Tab.5}
\end{center}
\end{table*}

\subsubsection{Evaluation on S3DIS}
 We perform a performance evaluation of various 3D point cloud semantic segmentation methods on the S3DIS dataset, and the comparison results are provided in Table~\ref{Tab.2}. From Table~\ref{Tab.2}, it can be seen that our method achieves state-of-the-art performance using only scene-level label supervision, surpassing the previous best method 3DSS-VLG by a margin of 6.5\%. Furthermore, our method also outperforms some fully supervised methods. These results collectively validate the effectiveness and superiority of our proposed methodology in leveraging weak supervision while maintaining competitive precision.

\subsection{Ablation}
\begin{table}[!t]
\centering
\begin{minipage}[t]{0.2\textwidth} % 左侧表格
\centering
\captionsetup{width=\linewidth} % 设置标题宽度与表格一致
\caption{\small Ablation studies of components on ScanNet dataset.}
\setlength{\tabcolsep}{1mm}{

\begin{tabular}{cccc}
 \toprule%\hline %绘制一条水平横线
    & CALR       & GALR        & mIoU \\
\midrule %绘制一条水平横线
(a) &            &             & 49.4 \\
(b) & \checkmark &             & 60.0 \\
(c) &            & \checkmark  & 51.5 \\
(d) & \checkmark & \checkmark  & \textbf{61.4} \\
\bottomrule %绘制一条水平横线
\end{tabular}}

\label{Tab.6}
\end{minipage}
\hfill % 填充空白，使两个表格分别位于左右两侧
\begin{minipage}[t]{0.24\textwidth} % 右侧表格
\centering
\captionsetup{width=\linewidth} % 设置标题宽度与表格一致
\caption{\small Performance with different $T$ in the STLP procedure on ScanNet dataset.}
\setlength{\tabcolsep}{1.4mm}{

\begin{tabular}{ccc}
\toprule
$T$  & mIoU  & mAcc  \\ \midrule
0 & 61.4 & 71.9 \\
1 & 62.8 & 72.7 \\
2 & \textbf{64.1} & \textbf{73.8} \\
3 & 63.7 & 73.6 \\ \bottomrule
\end{tabular}}
\label{Tab.7}
\end{minipage}
\end{table}

\subsubsection{Effectiveness of Each Component}
\label{ablation}
To explore the effectiveness of individual components in our proposed method, we conduct comprehensive ablation studies on the ScanNet dataset, with quantitative results presented in Table~\ref{Tab.6}. Ablation model (a) retains only the MinkowskiNet18A UNet~\cite{choy20194d} backbone and is trained directly using pseudo labels generated by retaining the top-$V$\% of points based solely on confidence scores. The cross-entropy loss is introduced to supervised this procedure. In contrast, model (b) utilizes only the CALR strategy to generate the pseudo labels, which selects the top-$V$\% within each category. Firstly, we analyze the pseudo labels between model (a) and model (b), and the visual comparison of the pseudo labels is depicted in Fig.~\ref{fig4}. From Fig.~\ref{fig4}, it is clearly evident that the pseudo labels generated by the model using CALR strategy are more accurate than the model without CALR strategy.

Additionally, we also provide the class-wise segmentation performance of the pseudo labels (Train) and model predictions (Val) on the ScanNet dataset in Table~\ref{Tab.5}. From this table, we can observe that selecting the top-$V$\% of points based on confidence scores introduces a significant class imbalance issue in pseudo label generation. In particular, the model tends to underrepresent or entirely omit small or rare object categories (\textit{e.g.}, shower curtains), thereby degrading segmentation quality. In contrast, our proposed CALR strategy effectively alleviates this problem by simultaneously improving pseudo-label accuracy and balancing their category distribution. By promoting a more balanced representation across classes, CALR helps the model learn more reliable and fine-grained semantic features, ultimately enhancing 3D segmentation performance. Furthermore, as demonstrated in Table~\ref{Tab.6}, the performance comparison between model (a) and model (b) reveals a substantial improvement in mIoU from 49.4\% to 60.0\%. These results confirm that our proposed CALR strategy generates more class-balanced and accurate pseudo labels, leading to superior segmentation outcomes.

Furthermore, we also conduct an ablation study to investigate the effectiveness of the GALR strategy, with quantitative and qualitative results presented in Table~\ref{Tab.6} and Fig.~\ref{fig4}, respectively. In Table~\ref{Tab.6}, model (c) corresponds to model (a) augmented with the GALR strategy, while model (d) represents model (b) enhanced by the GALR strategy. Notably, model (d) is supervised using pseudo labels initialized according to the methodology described in Sec.~\ref{plr}. Analysis of Table~\ref{Tab.6} reveals that model (c) achieves a 2.1\% performance improvement over model (a), and model (d) demonstrates a 1.4\% enhancement compared to model (b). In addition, from Fig.~\ref{fig4}, we can see that the pseudo labels generated by the model using the GALR strategy are closer to the ground truth and have clearer contours than the model without the GALR strategy. Besides, we perform an ablation study of the GALR strategy during the inference phase, with results detailed in Table~\ref{Tab.8}. These findings conclusively demonstrate that integrating 3D geometric priors improves pseudo-label quality, thereby enhancing the model's segmentation performance.

\begin{table}[t]
\caption{\small Ablation studies of GALR during inference.}
\begin{center}
\small
\begin{tabular}{cccc}
\toprule
Method  & ScanNet(OpenSeg)  & ScanNet(LSeg) &S3DIS \\ \midrule
$w/o$ GALR &55.7  &62.1  &51.2 \\
$w$ GALR & \textbf{57.5} &\textbf{64.1}  &\textbf{51.8} \\ \bottomrule
\end{tabular}
\label{Tab.8}
\end{center}
\end{table}
\subsubsection{Investigating the Influence of Top-$V$\% in CALR Strategy}
\begin{figure}[t]
\centering
\includegraphics[width=0.9\linewidth]{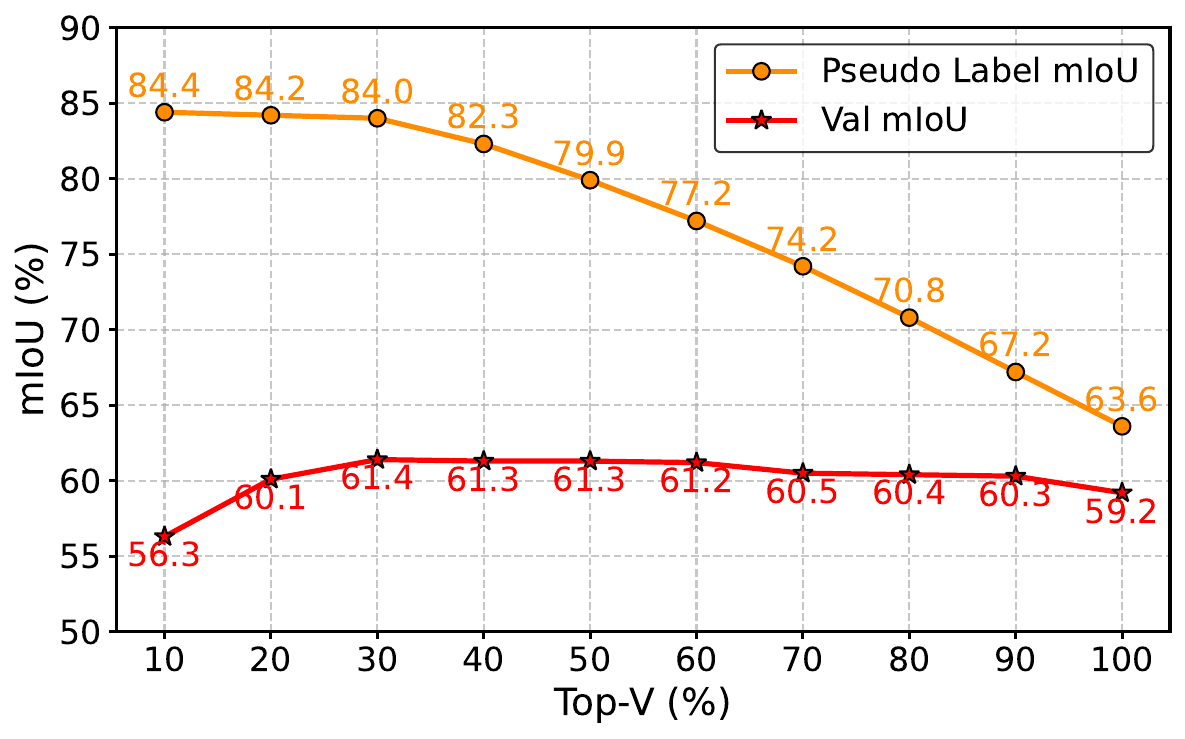}
\caption{\small A quantitative comparison about pseudo labels and predictions with different $V$ in the CALR strategy on ScanNet dataset.} 
\label{fig6}
\end{figure}

For sake of investigating the impact of retaining different proportions of pseudo labels (top-$V$\%) in the CALR strategy, we conduct comprehensive ablation experiments on the ScanNet dataset, and the comparison results are illustrated in Fig.~\ref{fig6}. From Fig.~\ref{fig6}, we can find that as the rate increases, the mIoU of pseudo labels on the training dataset decreases. This indicates that higher confidence thresholds in the CALR strategy correlate with improved pseudo-label quality. Regarding segmentation performance under varying top-$V$\% settings, the mIoU initially rises with the supervision rate but begins to decline after exceeding 30\%. When the retention rate of pseudo labels is low, a large proportion of points remain unlabeled, hindering the model's ability to perceive local scene details. Conversely, when the top-$V$\% exceeds a certain threshold, the degradation of pseudo-label quality introduces more noise, which disrupts model training and reduces performance. This suggests a trade-off between maintaining sufficient pseudo labels for effective guidance and preserving their quality. Therefore, to balance these factors, we select 30\% as the optimal retention rate for the CALR strategy.

\subsubsection{Investigating the Influence of the Overleap Threshold $\alpha$  in GALR Strategy}
\begin{figure}[t]
\centering
\includegraphics[width=0.9\linewidth]{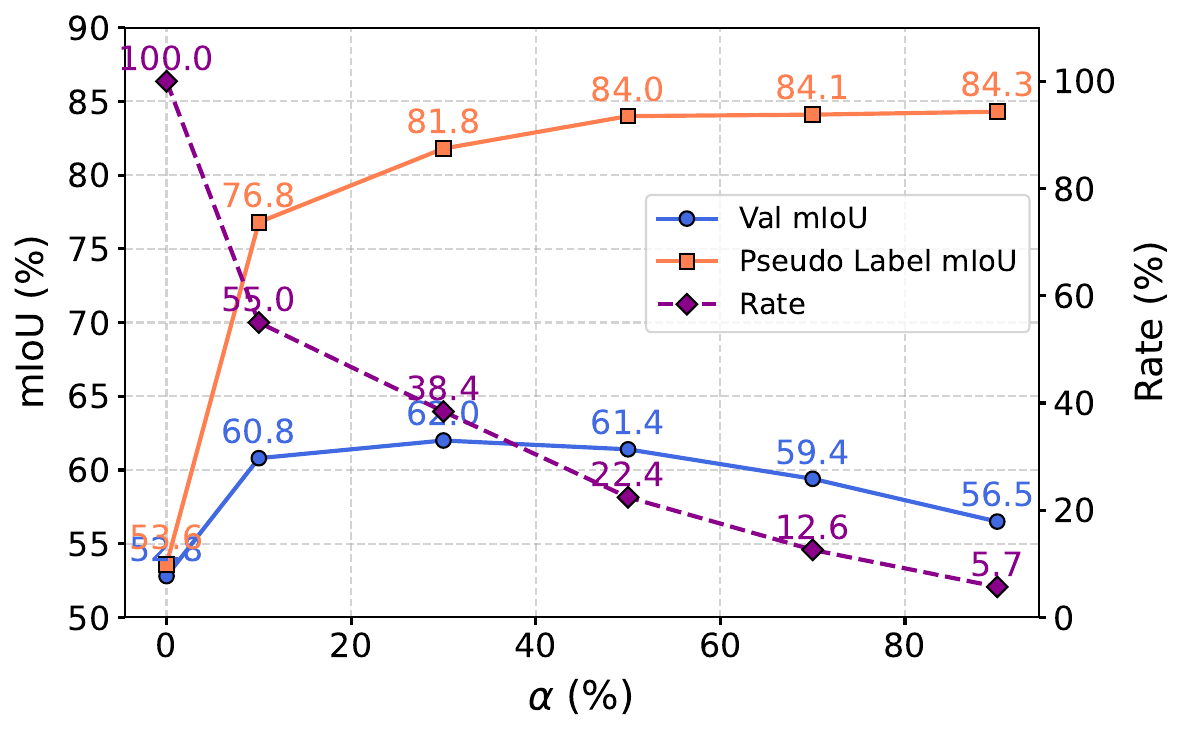}
\caption{\small A quantitative comparison about overleap threshold with different $\alpha$ in the GALR strategy on ScanNet dataset.} 
\label{fig7}
\end{figure}

We first analyze the relationship between pseudo-label performance and the labeled rate, and the mIoU curves are depicted in Fig.~\ref{fig7}. From Fig.~\ref{fig7}, we can observe that as the confidence threshold increases, the quality of pseudo labels improves while the labeled rate decreases. Regarding validation set performance, we observe that the model's performance initially increases with a rising threshold but subsequently declines. This trend arises because a higher threshold enhances pseudo-label quality, providing more accurate supervision for segmentation tasks. However, when the threshold becomes excessively large, the labeled rate drops significantly, resulting in insufficient pseudo labels to guide the model effectively, which ultimately leads to a decline in segmentation performance.

\subsubsection{Investigating the Influence of $T$ in STLP Procedure}
\label{infk}
\begin{figure*}[t]
\centering
\includegraphics[width=1.0\linewidth]{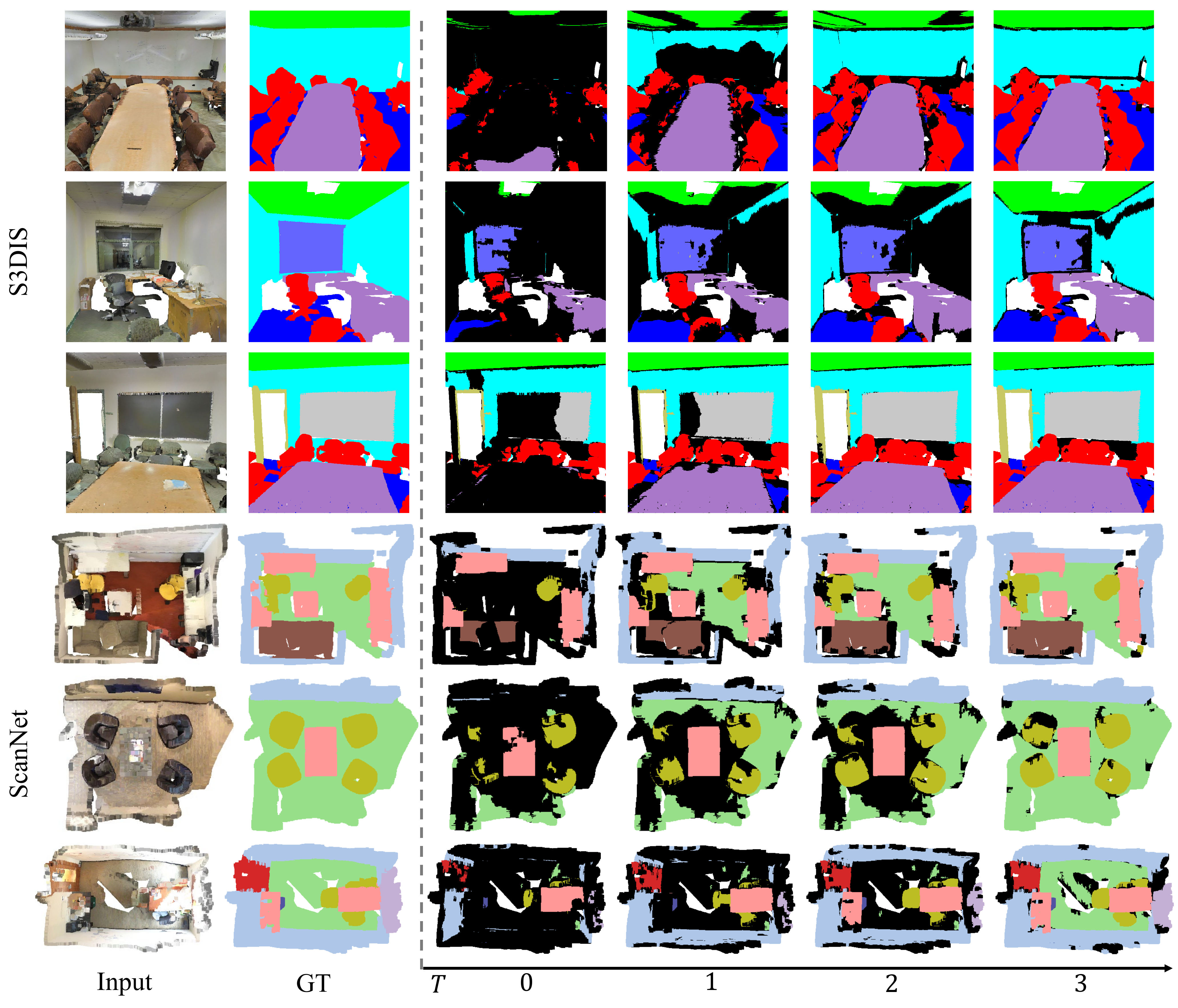} 
\caption{\small Progressive visualization of the behavior of pseudo labels throughout STLP. From left to right: input point clouds, ground truth and the subsequent pseudo labels updated at different timestep $T$.} 
\label{fig10}
\end{figure*}

We further investigate the impact of varying $T$ values in the STLP procedure using four distinct $T$ settings: $T \in \{0,1,2,3\}$. The experimental results are provided in Table~\ref{Tab.7}.  From this table, we find that increasing the number of self-training iterations enhances model performance, indicating that additional rounds of self-training facilitate label propagation to unlabeled regions. This process enriches the model with more semantic information, thereby benefiting 3D weakly supervised semantic segmentation. We adopt $T = 2$ as the default parameter, as excessively large $T$ values may lead to error accumulation in pseudo labels, which negatively impacts generalization. In addition, we also provide a progressive visualization of the behavior of pseudo labels throughout STLP in Fig.~\ref{fig10}. As shown in Fig.~\ref{fig10}, it can be clearly observed that the progressive refinements of pseudo labels during the STLP process are driven by the Label Update and GALR strategies. This observation indirectly validates the effectiveness of our proposed STLP module.

\begin{figure}[t]
\centering
\includegraphics[width=1.0\linewidth]{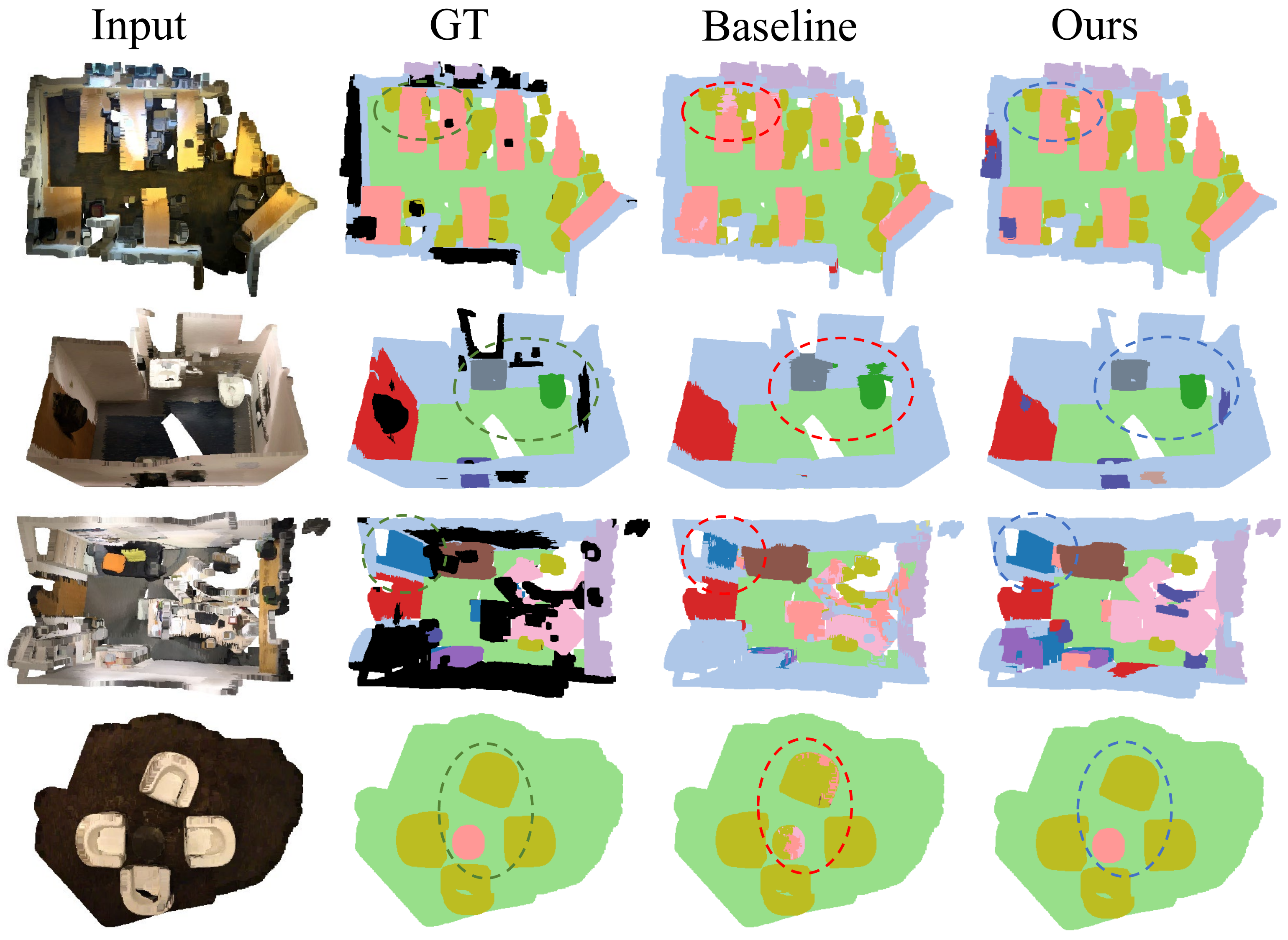} % Reduce the figure size so that it is slightly narrower than the column.
\caption{\small Qualitative results on the ScanNet dataset of baseline and our framework. From left to right: input point clouds, ground truth, baseline results, and our results.}
\label{fig8}
\end{figure}

\begin{table}[!t]
\centering
\begin{minipage}[t]{0.24\textwidth} % 左侧表格
\centering
\captionsetup{width=\linewidth} % 设置标题宽度与表格一致

\caption{\small  A quantitative comparison  with different pseudo labels update strategy in STLP.}

\setlength{\tabcolsep}{1.4mm}{
\begin{tabular}{cccc}
\toprule
$T$ & Full & Retained   & GT  \\ \midrule
0 & 61.4    & 61.4 & 65.3   \\
1 & 62.1    & 62.8 &  67.2   \\
2 & \textbf{63.5}    & \textbf{64.1} & 68.3   \\
3 & 62.8    & 63.7 &  \textbf{68.7}   \\ \bottomrule
\end{tabular}

}

\label{Tab.9}
\end{minipage}
\hfill % 填充空白，使两个表格分别位于左右两侧
\begin{minipage}[t]{0.23\textwidth} % 右侧表格
\centering
\captionsetup{width=\linewidth} % 设置标题宽度与表格一致
\caption{\small A quantitative comparison  with different top-$V$ and $\alpha$ on S3DIS dataset.}
\setlength{\tabcolsep}{1.2mm}{
\small
\begin{tabular}{ccc}
\toprule
   & $V$    & $\alpha$ \\ \midrule
10 & 43.6 & 46.9  \\
30 & \textbf{48.8} & 47.8  \\
50 & 48.3 & \textbf{48.8}  \\
70 & 47.1 & 47.4  \\
90 & 45.8 & 45.7  \\ \bottomrule
\end{tabular}}
\label{Tab.10}
\end{minipage}

\end{table}

\subsubsection{Investigating Pseudo Label Error Accumulation in STLP} In the STLP framework, pseudo labels generated in previous iterations are retained and reused in subsequent training steps. This design raises a potential concern about error accumulation, where incorrect labels might propagate across iterations and degrade segmentation performance. To address this, we conduct an ablation study comparing two strategies: (1) Full Update, which regenerates all pseudo labels at each iteration without retaining prior labels, and (2) Retained Update, which preserves and refines previously generated pseudo labels as described in Section III-C. The comparison results are provided in Table~\ref{Tab.9}. From this table, we observe that the performance of the Retained Update strategy is comparable to that of the Full Update, suggesting that error accumulation in STLP is minimal and well-controlled. To further explore the upper bound of STLP, we perform an oracle experiment where pseudo labels are replaced with ground-truth labels at every iteration. As shown in Table~\ref{Tab.9}, the marginal performance gain over our method underscores the high quality of generated pseudo labels and confirms that error propagation remains negligible throughout the training process. These findings validate the robustness of our label propagation mechanism in maintaining accurate supervision signals across iterations.

\subsubsection{Investigating the Computational Cost} To assess the computational efficiency of our proposed method, we conduct a detailed runtime analysis on the ScanNet dataset. The analysis includes two main components: (1) Data Preparation Time: Before training, we compute feature embeddings from multi-view images for each room. This step requires approximately 695 seconds per room. While embedding extraction is time-intensive, it can be efficiently managed through offline preprocessing prior to training, eliminating runtime overhead. (2) Computational Complexity: Our model is trained on an NVIDIA V100 GPU. For each iteration, the label propagation step consumes 260 seconds, and the entire training process requires approximately 17 hours. Although the total training duration is substantial, it remains justified given the model's performance and the scale of the dataset. These analyses provide practical insights for deploying our proposed method in resource-constrained environments.

\subsubsection{Investigating the Hyperparameter Setting} To systematically analyze hyperparameter sensitivity, in addition to conducting ablation experiments on ScanNet shown in Fig.~\ref{fig6} and Fig.~\ref{fig7}, we also conduct additional ablation studies on the S3DIS dataset and the results are shown in Table~\ref{Tab.10}, focusing on two critical parameters: the top-$V$ selection threshold and confidence threshold $\alpha$. By observing the results in Fig.~\ref{fig6}, Fig.~\ref{fig7}, and Table~\ref{Tab.10}, we find a consistent trend that setting the top-$V$ to 30\% and $\alpha$ to 0.5 can achieve the best trade-off between the quality and quantity of pseudo labels, ensuring reliable supervision signals while maintaining sufficient labeled data coverage for model training.

\begin{table}[t]
\centering
\begin{minipage}[t]{0.20\textwidth} % 左侧表格
\centering
\captionsetup{width=\linewidth} % 设置标题宽度与表格一致

\caption{\small Performance comparisons with different 3D backbones on ScanNet dataset.}
\setlength{\tabcolsep}{1.2mm}{
\begin{tabular}{ccc}
\toprule
Backbone  & mIoU  & mAcc  \\ \midrule
Mink14A & 60.9 & 71.3 \\
Mink18A & \textbf{61.4} & \textbf{71.9} \\
Mink34A & 61.3 & 71.9 \\ \bottomrule
\end{tabular}}
\label{Tab.11}
\end{minipage}
\hfill % 填充空白，使两个表格分别位于左右两侧
\begin{minipage}[t]{0.25\textwidth} % 右侧表格
\centering
\captionsetup{width=\linewidth} % 设置标题宽度与表格一致
\caption{\small Performance comparisons on ScanNet dataset with unsupervised methods.}
\setlength{\tabcolsep}{1.2mm}{
\small
\begin{tabular}{ccc}
\toprule
Method        & mIoU  & mAcc  \\ \midrule
GrowSP~\cite{zhang2023growsp} & 25.4 & 44.2 \\
U3SD$^3$~\cite{liu2024u3ds3} & 27.3 & 46.8 \\
CLIP-FO3D~\cite{zhang2023clip} & 30.2 & 49.1 \\ 
OpenScene~\cite{peng2023openscene} & 54.2 & 66.6 \\ 
Ours & \textbf{56.7} & \textbf{68.9} \\ 
\bottomrule
\end{tabular}}
\label{Tab.12}
\end{minipage}
\end{table}
\subsubsection{Extend to Unsupervised 3D Semantic Segmentation}
We also extend our proposed method to an unsupervised paradigm, where we cease using scene-level labels for filtering during both the Pseudo Label Generation and STLP procedures, while maintaining identical configurations for all other components. Experimental evaluations on the ScanNet dataset presented in Table~\ref{Tab.12} reveal that our approach achieves promising performance. This indicates that our devised CALR and GALR strategies are more effective in refining pseudo labels. Furthermore, the Label Propagation mechanism indicates strong scalability by effectively propagating pseudo labels to unlabeled regions, collectively validating the robustness and adaptability of our approach across different supervision paradigms.

\subsubsection{Experiments with Different Backbones} We report a performance comparison of our proposed method during the self-training procedure on the ScanNet dataset using different 3D backbones in Table~\ref{Tab.11}. Finally, follwing the previous work~\cite{xu20243d}, we also use the MinkowskiNet18A as our 3D backbone.
% From this table, it is clearly seen that using MinkowskiNet18A as our default 3D backbone network obtains better performance than other backbone settings. This selection ensures compatibility with existing benchmarks while leveraging the model's proven effectiveness in handling sparse 3D data. 

\subsubsection{Qualitative Results}
\begin{figure}[t]
\centering
\includegraphics[width=1.0\linewidth]{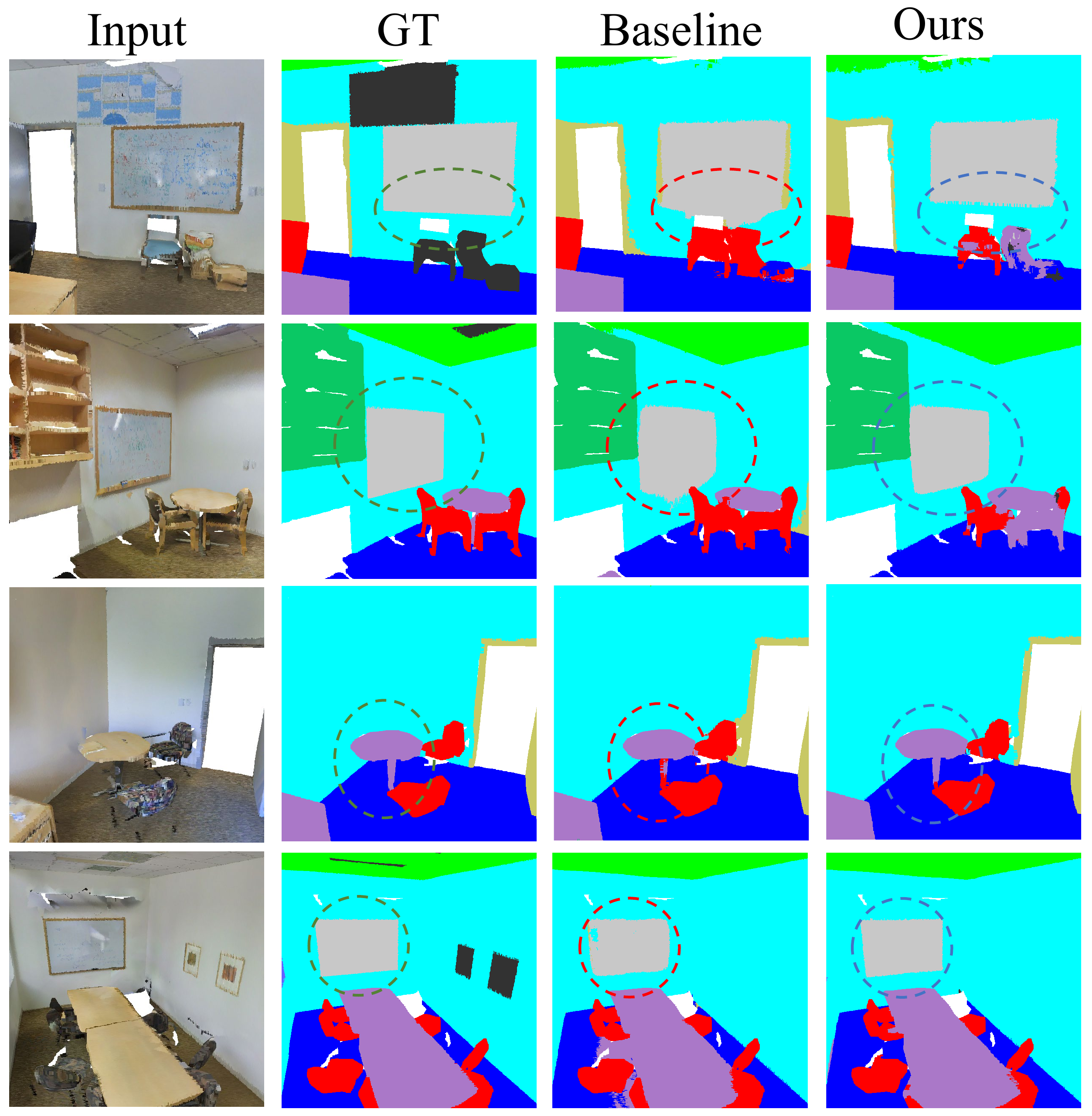} 
\caption{\small Qualitative results on the S3DIS dataset of baseline and our framework. From left to right: input point clouds, ground truth, baseline results, and our results.}
\label{fig9}
\end{figure}

We visualize the qualitative comparison of the proposed method and the baseline in Fig.~\ref{fig8} and Fig.~\ref{fig9}. The baseline corresponds to model (a) described in Sec.~\ref{ablation}. Compared to the results of the model (a) and Ours, we can see that the results generated by Ours are closer to the ground truths than the model (a). Notably, our approach excels in handling objects with complex geometric boundaries, achieving more precise contour delineation. This enhancement is primarily attributed to the GALR strategy, which integrates 3D geometric knowledge into pseudo-label generation, thereby guiding the model to better capture spatial relationships. Additionally, our designed CALR and Label Propagation strategies work synergistically to refine pseudo labels, enabling more accurate segmentation results across various object categories.

\subsection{Limitations}
Our method still relies on scene-level labels as filter masks, yet effectively utilizing them to guide the model in perceiving scene categories remains a challenging problem. Furthermore, our incorporation of 3D geometric priors is currently limited to indirectly leveraging superpoint information. Exploring ways to directly integrate 3D geometric knowledge into the model constitutes an important avenue for future research.

\section{Conclusion}
In this paper, we propose a simple yet efficient 3D weakly supervised semantic segmentation approach that integrates 3D geometric priors with class-aware semantic segmentation. In particular, our approach employs the Class-Aware Label Refinement module to generate more class-balanced and accurate pseudo labels, while the Geometry-Aware Label Refinement module is utilized to implicitly incorporate 3D geometric cues for further label refinement. Moreover, we design a self-training procedure to propagate pseudo labels to unlabeled regions, effectively enhancing segmentation quality through iterative optimization. Comprehensive experiments demonstrate that our proposed method significantly outperforms previous state-of-the-art approaches. Significantly, when adapted to an unsupervised learning paradigm, our method maintains promising performance, further substantiating its robustness and generalizability. While certain limitations persist, particularly in handling complex scenes with severe class imbalances, our work highlights the potential of leveraging scene-level labels and 3D geometric priors as a promising avenue for future research in 3D semantic segmentation.

\bibliographystyle{IEEEtran}
\bibliography{IEEEtran}
\vfill
\end{document}